\documentclass[journal,transmag]{IEEEtran}
\usepackage{graphicx}
\usepackage{tabularx}
\usepackage{amsmath}
\usepackage{subfloat}
\usepackage{underscore}
\usepackage{url}
\usepackage{algorithm}
\usepackage{algpseudocode}
\usepackage{textcomp}
\usepackage{amssymb}
\usepackage{cite}
\usepackage{epstopdf}
\usepackage{subfigure}
\begin{document}
\UseRawInputEncoding
\title{A Bayesian Deep Learning Technique with an Advanced Divergence Scheme for Multi-Step Ahead Solar Generation Forecasting}
\author{\IEEEauthorblockN{Deakin University, Australia}}
\author{Devinder Kaur, Shama Naz Islam, and Md. Apel Mahmud
	\thanks{D. Kaur, S. N. Islam and M. A. Mahmud are with School of Engineering, Deakin University, Australia. emails: devinderkaur@deakin.edu.au, shama.i@deakin.edu.au, apel.mahmud@deakin.edu.au.
	
	A shorter version of this paper has been presented in STPEC 2021.}}
\maketitle
\IEEEdisplaynontitleabstractindextext
\IEEEpeerreviewmaketitle
\textbf{Abstract} - 
In this paper, we propose an improved Bayesian bidirectional long-short term memory (BiLSTM)-based deep learning technique for multi-step ahead (MSA) solar generation forecasting.
The proposed probabilistic technique applies alpha-beta divergence for a more appropriate consideration of outliers in solar generation data and resulting variability of the weight parameter distribution in Bayesian neural networks.
The proposed method is examined on highly granular solar generation data from Ausgrid using probabilistic evaluation metrics such as Pinball loss and Winkler score. 
Moreover, 
a comparative analysis is provided
to test the effectiveness of the improved MSA forecasting method for variable forecasting horizons. The numerical results clearly demonstrate that the improved Bayesian BiLSTM with alpha-beta divergence outperforms standard Bayesian BiLSTM and other benchmark methods for MSA forecasting in terms of error performance.

\section{Introduction}
With the rapid increase in carbon emissions, there is a dire need of clean and sustainable energy systems in the world today. 
In this regard, solar power is one of the fastest growing renewable energy sources (RES) penetrating into the power grid around the globe \cite{solarstats2021}.
With the substantial deployment of smart meters in the modern power grid, a huge amount of data related to energy generation and electricity demand can be acquired \cite{wang2018review}. This spatio-temporal energy data can be utilized by forecasting methods to predict future energy demand and generation  to support grid operations \cite{kaur2020energy}. 

Over the past decades, statistical and machine learning (ML) methods such as 
autoregressive \cite{messner2019online} and support vector regression (SVR) \cite{ghelardoni2013energy} respectively, have been widely utilized
to forecast future energy trends related to customer electricity demand \cite{saeedi2021adaptive} and renewable power generation involving rooftop solar panels \cite{Cyril:2017}. 
While majority of existing literature in energy forecasting
involves single-step ahead (SSA) predictions over short time horizons, it is not sufficient for the long-term planning applications such as generation scheduling where bids need to be submitted in a day-ahead manner \cite{parlos2000multi}. 
Moreover, having multiple values predicted in long- and short-term horizons, the variability and projected abnormality can be analyzed more efficiently  \cite{cheng2006multistep}.
Thus, it is crucial to forecast
multi-step ahead (MSA) energy generation with
longer time horizons
\cite{bontempi2014monte}.

Main challenges of MSA forecasting include accumulated errors and increased uncertainty in the predictions, contrary to SSA point forecasting \cite{taieb2012review}.
To achieve the desired accuracy, different MSA approaches need to be contemplated in addition to a robust forecasting framework. 
In this regard, Taieb \textit{et al.} presented a comparative study of various MSA approaches namely direct, recursive, multi-input multi-output (MIMO), and direct MIMO (DIRMO) \cite{taieb2009long}. Among these approaches, DIRMO has a unique advantage in retaining model flexibility and avoiding accumulated errors, which will be an important factor for deep learning (DL)-based MSA forecasting techniques.
Similarly, Wang \textit{et al.} \cite{wang2016analysis} investigated aforementioned MSA approaches to make wind speed predictions over longer time horizons and 
observed better accuracy with the DIRMO approach.

Furthermore, 
direct, MIMO, and recursive approaches are inspected in \cite{bao2014multi} using SVR, where MIMO-SVR is found to have the least mean absolute percentage error (MAPE) despite a computational load trade-off.
In a similar manner, \cite{peng2020multi} inspected a delay-embedding-based ML scheme for solar irradiance and wind speed forecasting for $100$ and $150$ time-steps ahead, respectively. However, the presented scheme relies on the availability of multi-dimensional spatial-temporal data which may not always be accessible.
In this context, DL methods can manage the applications with data insufficiency more effectively \cite{shi2017deep}.

Notably, artificial neural networks (ANN)-based DL techniques such as recurrent neural networks (RNN) have set a performance benchmark for energy forecasting applications, as they can deal with temporal and nonlinear features of energy datasets very efficiently over different time horizons \cite{kaur2019smart}.
In this context, Ahmed \textit{et al.} presented a six-steps (half hourly with an interval of $5$ min.) ahead wind forecasting scheme using autoregressive neural networks using direct and recursive methodologies
\cite{ahmed2017multi}, however the scheme could not be extended to forecasting horizons greater than six-time steps.
Recently, authors in \cite{mellit2021deep} have compared various DL methods such as RNN, long-short term memory  (LSTM), gated recurrent unit (GRU), and convolutional neural networks (CNN) for solar photovoltaic (PV) power forecasting in both SSA and MSA manner and claimed LSTM to be best performing DL model amongst all. 
Furthermore, BiLSTM as a bidirectional component of LSTM serves as 
a more effective model due to its ability of back-and-forth learning
\cite{kaur2021vae}.
In this regard, Gairaa \textit{et al.} also utilized multiple linear regression (MLR) and DL-based MSA approach for  six-hours ahead solar irradiance prediction but the accuracy has not been investigated for sites with high variability \cite{gairaa2021contribution}.

Although
LSTM and BiLSTM provide superior performance for both SSA and MSA predictions \cite{toubeau2018deep,jahangir2020deep} as compared to other state-of-the-art ML methods, the accuracy for MSA-DL frameworks can be further improved and the consideration of longer time horizons also remains an open challenge \cite{taieb2012review,taieb2009long,wang2016analysis}.
In addition, 
standard DL-based forecasting techniques are deterministic in nature and suffer from uncertainty in model parameters (weights) and prediction outputs \cite{blundell2015weight}, which become significant when multiple time-steps are considered \cite{gal2016uncertainty}. In this context, probabilistic approaches have been widely used as effective methods to manage uncertainties in the renewable generation data \cite{Illindala2018}. 
Thus, there is a scope for an improved DL algorithm incorporated with probabilistic methods for effective MSA generation forecasting, which can tackle the uncertainty challenges of RESs and weight parameters efficiently.

In this regard, Bayesian inference integrated with neural layers \cite{yang2019bayesian}, specifically LSTM \cite{sun2019using} and BiLSTM \cite{kaur2021vae} has emerged as a potential solution to deal with the problem of weight and prediction uncertainty in the energy forecasting domain. 
Contrary to the standard DL models, these methods can generate prediction intervals (PI), which can provide lower and upper bounds for the future prediction values with a given probability. Our previous work \cite{kaur2021msa} has considered Bayesian DL (BDL) for MSA predictions which can be highly useful for renewable energy forecasting applications such as solar generation forecasting. 
However, existing research on BDL forecasting relies on standard variational inference (VI) techniques based on Kullback-Leibler (KL) divergence. Standard BDL methods underestimate the variance of weight parameter distribution and exhibit poorer performance for datasets having outliers \cite{regli2018alpha}, which can be a critical factor when there is high variability in the renewable energy data and in addition, MSA forecasting is performed. As a result, the standard VI techniques used in \cite{kaur2021msa} are not sufficient to ensure the required accuracy for MSA solar generation forecasting and an improved approach needs to be developed which can quantify uncertainty with more sharpness and calibration.

As a result, we make following contributions in this paper:
\begin{itemize}
	\item An improved Bayesian probabilistic method based on alpha-beta divergence integrated with Bidirectional LSTM (BiLSTM) is developed to approximate posterior distribution of weight parameters in the presence of uncertainties. The $improved$-Bayesian BiLSTM can quantify 
	uncertainties 
	and provide probabilistic forecasts using PIs
	more accurately. 
	
	\item The proposed Bayesian method is implemented to generate MSA forecasts using DIRMO approach for one day ahead solar generation, i.e., 48-time steps ahead (half hourly). Furthermore, the developed method has also been tested for different forecasting horizons and it is evident from the results that it can manage longer forecasting horizons more effectively.
	
	\item A comparative case study is presented using Ausgrid solar generation dataset to demonstrate the superior performance of the proposed MSA forecasting framework against the benchmark SSA and MSA forecasting methods. 
\end{itemize}

The rest of the paper is organized in the following manner.
Section \ref{sec:problem} describes the problem formulation for optimizing the BDL model using BiLSTM in the presence of parameter uncertainties. Section \ref{sec:Proposed} outlines the improved Bayesian BiLSTM approach integrated with direct MIMO (DIRMO) for MSA solar generation forecasting. Section \ref{sec:results} highlights the implementation results. Finally, Section \ref{sec:conclusion} concludes the paper.
\begin{figure}[!t]
	\centering
	\includegraphics[width=0.425\textwidth]{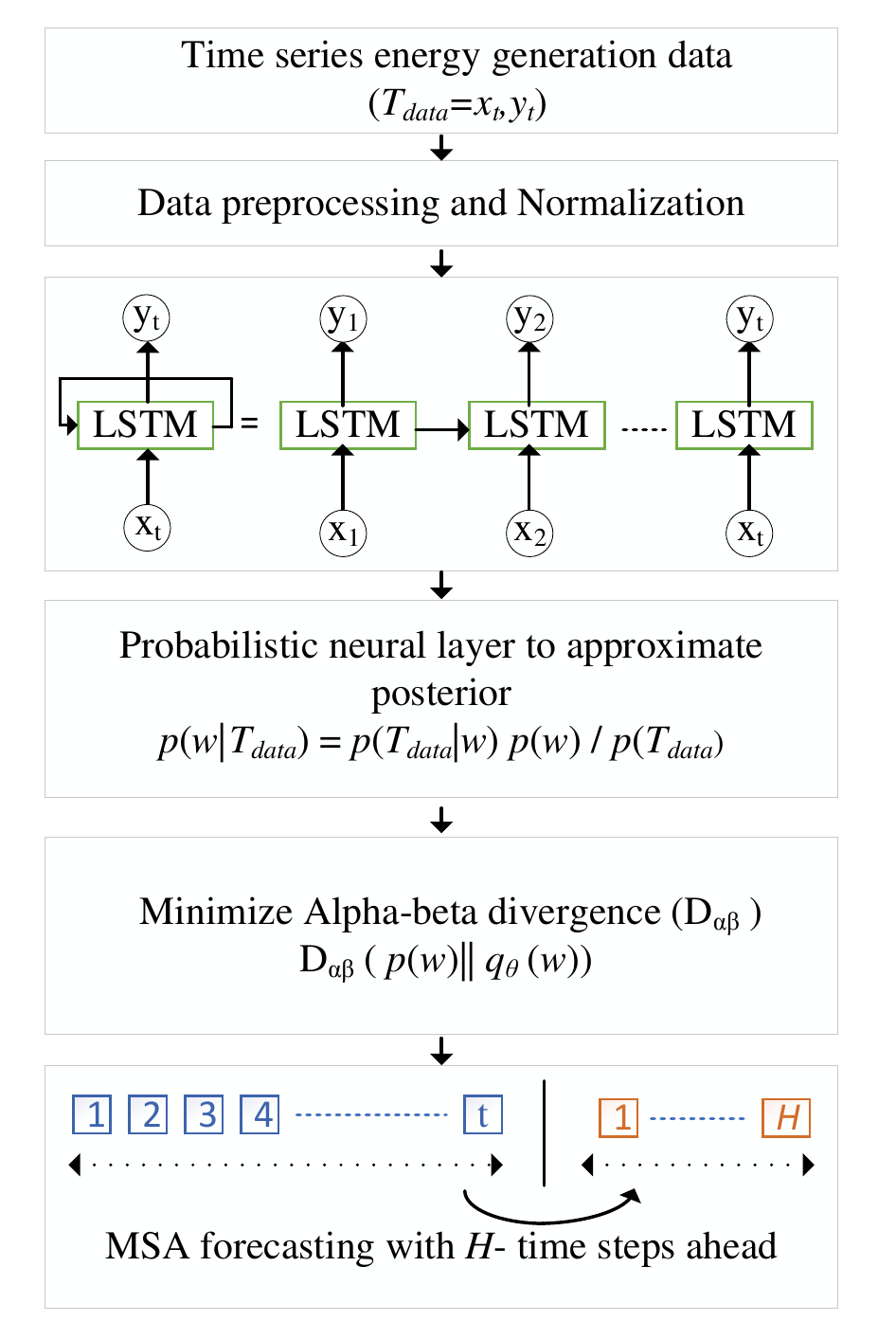}
	\caption{Flowchart of the proposed $Improved$-Bayesian BiLSTM}
	\label{flowchart}
\end{figure}
\section{preliminaries and Problem Formulation}\label{sec:problem}
In this section, we discuss preliminaries on BiLSTM architecture and formulate the parameter optimization problem for BDL model considering uncertainties in model parameters and solar generation.
\subsection{Preliminaries on Bidirectional LSTM}
This subsection briefly outlines the architecture of one LSTM cell employed in a bidirectional setting (BiLSTM) to process time series solar generation data with an intensive learning approach for improved MSA forecasting.
Each LSTM cell involves four layers (cell state and three gates) in addition to RNN's recurrent loop structure. 
The three types of gates namely input ($i_{t}$), output ($o_{t}$), and forget ($f_{t}$) gates enable the network to remove or add the information at $t$ time stamp via cell state ($c_t$) with the help of $sigmoid$ function \cite{colah}.
The equations for gating mechanism \cite{hochreiter1997long} are given as:

\begin{equation}\label{eq:inputgate}
i_{t} =\sigma(w_{i}[y_{t-1}, x_{t}] + b_{i})
\end{equation}
\begin{equation}\label{eq:forgetgate}
f_{t}=\sigma(w_{f}[y_{t-1},x_{t}]+b_{f})
\end{equation}
\begin{equation}\label{eq:outputgate}
o_{t}=\sigma(w_{o}[y_{t-1},x_{t}]+b_{o})
\end{equation}
where $\sigma$ denotes the $sigmoid$ function. Furthermore, $w_{i}$, $w_{f}$, and $w_{o}$ are the weight matrices 
and $b_{i}$, $b_{o}$, $b_{f}$ denote the bias vectors corresponding to each gate.
For a current input 
$x_t$ 
the information at previous state ($y_{t-1}$) is assessed by forget gate and is either discarded (using $\sigma$) or updated (using $\sigma$ and tanh) to the input gate using a candidate cell state ($\tilde{c_{t}}$) as:
\begin{equation}\label{eq:cellprevious}
\tilde{c_{t}}=\tanh(w_{c}[y_{t-1},x_{t}]+b_{c})
\end{equation}
\begin{equation}\label{eq:cell}
c_{t}=(c_{t-1} * f_{t} ) + (i_{t} * \tilde{c_{t}})
\end{equation}
where $\tanh$ represents the hyperbolic tangent activation function. 
Note that, $w_c$ and $b_c$ refer to weight and bias corresponding to cell state.
The filtered output ($y_t$) for $x_t$ is obtained using:
\begin{equation}\label{eq:lstmoutput}
y_t = o_t * \tanh(c_t)
\end{equation} 

To train LSTM network, back-propagation through time (BPTT) is used as a training algorithm. 
The error derivatives are computed over weight parameters as a difference between predicted and estimated values. 
The total error ($\epsilon$) for total number of samples ($tot$) is calculated as:
\begin{equation}
\frac{d\epsilon}{dw} = \sum\limits_{t=1}^{tot} \frac{d\epsilon_{t}}{dw}
\end{equation}
The $\epsilon$ needs to be minimized as:
\begin{equation}\label{para}
\textrm{arg} \min_{\phi}{(\epsilon)}
\end{equation}
where $\phi$ represents the hyperparameters such as learning rate, dropout rate, train-test and validation split ratio, number of epochs, etc. to tune model parameters in a neural network.
\subsection{Problem formulation for Bayesian deep learning models}
The proposed probabilistic scheme utilizes the concept of Bayesian neural networks (BNN).
The idea is to learn the distributions of weight parameters instead of deterministically learning specific weights (and bias) in a deep neural network \cite{Zhang:2020}. The weight distributions are further used to sample the targeted output from the input data
\cite{izmailov2021bayesian}.
For a time series power generation dataset $T_{data} = (x_{t},y_{t})$, the posterior distribution of weight parameters in a BNN \cite{ho1964bayesian} is defined as:

\begin{equation}\label{eq:bayes}
p(w|T_{data})=\frac{p(T_{data}|w)p(w)}{p(T_{data})}
\end{equation}
where $y_t$ represents the actual solar generation data for the $t^{th}$ time interval. Further, $p(w|T_{data})$ is known as the posterior distribution function of weight parameters ($w$) and $p(w)$ is the prior distribution symbolizing the assumed probability of $w$ prior to any evidence is obtained. 
Using evidence $T_{data}$, prior is updated by multiplying likelihood $p(T_{data}|w)$ with $p(w)$ as given in the numerator of \eqref{eq:bayes}.

Note that computing posterior is the key element to obtain a full distribution for $w$ and it often represents the epistemic or weight uncertainty in Bayesian neural layers \cite{wilson2020bayesian}.
To compute full posterior, the product of likelihood and priors is evaluated over all parameters and further normalized using the technique of marginalization. The probability  of evidence ($p(T_{data})$) acts as a normalization constant used for marginalization and is computed using integration on each parameter setting as:
\begin{equation}\label{eq:marginalization}
p(T_{data})=\int p(T_{data}|w)p(w)dw
\end{equation}
Considering the large sample space generated using  \eqref{eq:marginalization}, computing the exact posterior analytically becomes impossible. To solve this, variational inference (VI) is utilized to obtain an approximated posterior distribution in the modern Bayesian deep learning domain \cite{kingma2013auto}. Thus, the network is trained using VI to learn the parameters of posterior distribution and optimized using negative log likelihood (NLL) cost function. 
Generally, the deviation between actual and approximated posterior is determined using divergence techniques \cite{zhang2018advances} such as 
KL divergence as:
\begin{equation}\label{eq:KL}
D(p||q)=  \log{p(T_{data})}- \mathop{\mathbb{E}}_{q_\theta(w)}[\log \frac{p(T_{data},w)}{q_\theta(w)}]
\end{equation}
where $q_{\theta}(w)$ denotes the approximated posterior over $\theta$ parameters using VI, signifying the mean and variance of the approximated posterior distribution.
Therefore, the objective function for the Bayesian deep learning model aims to minimize $D_{KL}$ between approximated and actual posterior distribution as shown below:
\begin{equation}\label{eq:opt}
\textrm{arg} \min_{q_\theta(w)} D(p||q)
\end{equation}


\section{Proposed MSA Solar Generation Forecasting Approach}\label{sec:Proposed}
In this section, we propose an improved Bayesian deep learning approach for MSA solar generation forecasting. Fig. \ref{flowchart} represents a conceptual flow chart of the proposed $Improved$-Bayesian BiLSTM method for MSA forecasting. The proposed scheme utilizes Bayesian BiLSTM model integrated with an improved divergence technique and applies DIRMO method for MSA forecasting.
MSA forecasting involves predicting a sequence of future half-hourly interval solar generation values in a horizon $H$ for a given time series with historical solar generation data. 
A time series $X_{data}$ for historical solar generation values at half-hourly intervals can be presented as:
\begin{equation}
X_{data}=[x_{1},x_{2},x_{3},.....,x_{t}]
\end{equation}

\noindent where $x_t$ is the solar generation at the $t^{th}$ interval. The SSA predictions (next half-hour solar generation) at the $(t+1)^{th}$ time interval is defined as:
\begin{equation}
\hat{y}_{t+1}=f(x_{1},x_{2},x_{3},.....,x_{t})
\end{equation}
Similarly,
\begin{equation}
\hat{y}_{t+2}=f(\hat{y}_{t+1})
\end{equation}
and so on.        
A segment $X_{data(t-k)}^{t}$ from the training dataset $X_{data}$, representing solar generation data from $(t-k)^{th}$ interval to $t^{th}$ interval for $k<t$, can be defined as:
{\fontsize{9}{10.8}\selectfont
\begin{equation}
 X_{data(t-k)}^{t}=[x_{t-k},x_{t-k+1},x_{t-k+2},.....,x_{t}]
\end{equation}}

\noindent Here $k$ represents the number of past observations of solar generation which are used to generate the MSA forecasts.
The function to obtain MSA forecasts for $h$-time steps ahead is formulated as:
{\fontsize{9}{10.8}\selectfont
\begin{equation}\label{eq:msa}
	[\hat{y}_{t+1}^{t+h}]=f(\hat{y}_{t}^{t+h-1},X_{data(t-k)}^{t}) \forall h \in [1,...,H]
\end{equation}}

\noindent where $\hat{y}_{t+1}^{t+h}$ ($\hat{y}_{t}^{t+h-1}$) are the solar generation output forecasts for the time intervals $(t+1)$ to $(t+h)$ ($t$ to $t+h-1$).
The proposed MSA probabilistic deep learning forecasting provides the distribution of multiple forecasts over a prediction horizon. So, the $T_{data}$ in \eqref{eq:bayes} becomes:
{\fontsize{9}{10.8}\selectfont
\begin{equation}
p(T_{data})=p(y_{t+1},y_{t+2},.....,y_{t+h}|x_{1},x_{2},....,x_{t})
\end{equation}}
Since the proposed methodology incorporates BiLSTM for $h$-time steps ahead forecasting, the objective function in \eqref{para} becomes:
{\fontsize{9}{10.8}\selectfont
\begin{equation}
\textrm{arg} \min_{\phi}{(\epsilon_{t+1}^{t+h})}
\end{equation}}
\noindent where
{\fontsize{9}{10.8}\selectfont
\begin{equation}
\epsilon_{t+1}^{t+h}=\sum_{t+1}^{t+h}\epsilon
\end{equation}}

To manage the uncertainties associated with optimizing the model parameters of the aforementioned MSA forecasting problem, a Bayesian deep learning approach is integrated with the BiLSTM model. In this approach, the prior probability distribution of the weight parameters is modeled as a standard normal distribution $\mathcal{N}(0,\sigma_i^2)$ with zero mean and standard deviation as $\sigma_i$. On the other hand, the posterior distribution is modeled by drawing random samples from $\mathcal{N}(0,\textbf{I})$ with zero mean and unity variance, shifting the samples by $\mu$ and then scaling by $\sigma_p$. Here $\mu$ and $\sigma_p$ represent the mean and standard deviation of the approximated posterior distribution of the weight parameters. $\sigma_p$ is modeled as a softplus function. Then the difference between true and approximated posterior is minimized using divergence techniques.

Standard BDL methods use KL divergence to optimize the posterior distribution of the weight parameters. However, KL divergence faces the challenge of underestimating the posterior variance \cite{minka2005divergence}, especially when the solar generation data has high variability. As a result, a more robust divergence technique which can manage the variability appropriately is required for optimizing the posterior distribution effectively. To address this issue, the proposed method integrates an improved divergence technique, namely, alpha-beta 
divergence ($D_{\alpha\beta}$) as a more scalable alternative of KL divergence to approximate posterior using parameters $\alpha$ and $\beta$. It gives more freedom for parameter selection according to the probability density and can be derived as:
{\fontsize{9}{10.8}\selectfont
\begin{align}\label{eq:finalalphabeta}
D_{\alpha\beta}(p||q) & = \left( \frac{\alpha+\beta-1}{\beta(\alpha+\beta)} - \frac{\alpha + \beta -1}{\alpha\beta}-\frac{1}{\alpha (\alpha+\beta)} + \frac{1}{\alpha} \right) \nonumber \\ & \times \mathbb{E}[\log (\theta)]
\end{align}}

The detailed derivation for $D_{\alpha\beta}$ is illustrated in the Appendix through {\eqref{eq:2}-\eqref{eq:5}}.
Thus, the weight parameter optimization problem transforms to:
{\fontsize{9}{10.8}\selectfont
\begin{equation}\label{eq:optproposed}
\textrm{arg} \min_{q_\theta(w)} D_{\alpha\beta}(p||q)
\end{equation}}
\begin{algorithm}[!h]
	{	\textbf{Input:} ${T_{data}} = (x_t, y_t)$  \\
		\textbf{Output:} $\hat{y}_{t}$,...,$\hat{y}_{t+h}$, $\hat{y}_{t,\tau}$, $\hat{y}_{t+h,\tau}$} for the $\tau^{th}$ percentile\\
		\textbf{Hyperparameters:} $\phi$\\
		\textbf{Parameters:} $w$\\
		\textbf{Performance metrics:} $Q_l$, $W_l$, $RMSE$, $MAE$
	\begin{algorithmic}[1]
		\State Obtain $T_{data}$ from half-hourly solar generation dataset;
		\State Scale and normalize $T_{data}$;
		\State $x_{t}$ $\implies$ explanatory variable ($2$D);
		\State $y_{t}$ $\implies$ observed variable ($2$D);
		\State Convert $x_{t}$ $\to$ $3$D;	
		\State Split $T_{data}$ into training and testing subsets;
		\State Follow steps \ref{st:startmodel}-\ref{st:endmodel} to train Bayesian inference;
		\For {($ep$=1; $ep$ $\leq$ $N$; $ep$++)}
		\State $p(w)\longleftarrow\mathcal{N}(0,\sigma_i^2)$;\label{st:startmodel}
		\State $q_\theta(w)\longleftarrow\sigma_p\odot\mathcal{N}(0,\textbf{I})+\mu$;
		\State Compute $D_{\alpha\beta}(p||q)$  using \eqref{eq:finalalphabeta};
		\State Minimize $D_{\alpha\beta}$ using \eqref{eq:optproposed};
		\State Input $x_{t}$ from training subset to BiLSTM layer;
		\State Calculate $i_t$, $f_t$, and $\tilde{c_{t}}$ using \eqref{eq:inputgate}, \eqref{eq:forgetgate} and \eqref{eq:cellprevious}, respectively;
		\State Update cell state using \eqref{eq:cell};
		\State Update $o_t$ using \eqref{eq:outputgate};
		\State Update output ($y_t$) of LSTM block using \eqref{eq:lstmoutput};
	    \State Pass the inputs in forward and backward directions and update $w$;\label{st:endmodel}
		\EndFor	
			\State Obtain the distribution of $t^{th}$ to $(t+h)^{th}$ predictions as output using \eqref{eq:msa};
			\State Compute the prediction intervals based on $\hat{y}_{t,\tau}$;
			\State Draw samples from the distributions of the forecasting model outputs;
			\State Compute the predictive mean of the generation forecast outputs;
			\State To evaluate probabilistic forecasting performance, follow steps \ref{st:ql} and \ref{st:wl} on test subset; 	
		\State Compute $Q_{l}$ using \eqref{pinball};\label{st:ql}
		\State Compute $W_{l}$ using \eqref{eq:winkler};\label{st:wl}
		\State To evaluate deterministic forecasting performance, follow steps \ref{st:rmse} and \ref{st:mae} on test subset; 
		\State Compute $RMSE$ using \eqref{rmse}; \label{st:rmse}
		\State Compute $MAE$ using \eqref{mape};\label{st:mae}	
\end{algorithmic} 
	\caption{An $Improved$-Bayesian BiLSTM technique for MSA solar generation forecasting} \label{algorithm:1}
\end{algorithm}
\begin{table*}[!h]
	\centering
	\renewcommand{\arraystretch}{1.5}
	\vspace{-20pt}
	\caption{Comparative analysis of proposed method for MSA solar Generation forecasting for ($t+48$)}\label{table:msa}
	\begin{tabular}{p{5mm}p{45mm}p{12mm}p{12mm}p{12mm}p{12mm}p{12mm}}
		\hline
		Sr. no. & Method &RMSE (kWh) &MAE (kWh) & Pinball\quad (avg) (kWh) &Winkler (kWh)& Time (s)\\
		\hline	
	1&	Standard BiLSTM (point forecasting)&$0.2634$&$0.1485$&-&-&$435.83$\\	
	2&	Standard LSTM (point forecasting)&$0.4041$&$0.2109$&-&-&$222.04$\\
	3&	Standard RNN (point forecasting)&$0.5621$&$0.4129$&-&-&$139.67$\\
	4&	Bayesian RNN \cite{yang2019bayesian}&$0.1422$&$0.1113$&$0.3431$&$3.1330$&$323.39$\\		
	5&	Bayesian LSTM \cite{sun2019using}&$0.1011$&$0.0535$&$0.2693$&$2.9820$&$204.56$\\
	6&	Bayesian BiLSTM \cite{kaur2021vae}&$0.0975$&$0.0486$&$0.2806$&$1.7400$&$412.33$\\
	7&	$Improved$-Bayesian BiLSTM&$ 0.0860$&$0.0387$&$0.0161$&$0.0143$&$174.37$\\
		\hline
	\end{tabular}
\end{table*}

Furthermore, Algorithm \ref{algorithm:1} presents the pseudo-code for the proposed MSA Bayesian BiLSTM forecasting approach. As illustrated, the obtained time series data ($T_{data}$) is preprocessed and normalized before dividing it into the training and testing subsets (lines 1-6). In addition, the two-dimensional (2D) explanatory variable ($x_t$) is converted into three-dimensional (3D) array representing historical generation data as time lags (samples, time-lags, features) (line 5). This 3D data from training set is then fed to the $improved$-Bayesian BiLSTM deep neural network to generate multi-time steps ahead predictions via output layer.
The proposed model is trained for $N$ iterations called epochs (lines 9-18). From probabilistic perspective, for each iteration, the posterior distribution is approximated for $w$ using variational inference by optimizing $D_{\alpha\beta}$ between prior and posterior distribution (lines 11-12). 

The training data is passed forward and backward in multiple slices through the network. The network adjusts weights based on the Bayesian inference. 
The model is trained until the value of the cost function cannot be further reduced. After achieving a global minima, the predictions for $t^{th}$ to $(t+h)^{th}$ intervals are computed (line 20). 
Note that $t+h$ number of neurons are passed in the output layer to generate multiple outputs at each step contributing towards DIRMO approach.
Then, predictive mean of the forecast outputs for all time intervals within horizon $H$ are computed (line 22). Moreover, prediction intervals (PIs) are generated using $\hat{y}_{t,\tau}$ for desired percentiles ($\tau$) values. 
Furthermore, to evaluate the forecasting performance in deterministic and probabilistic format, different error values are computed (lines 23-28). 
\section{Results and discussions}\label{sec:results}
This section includes the implementation results for the proposed technique carried out on solar generation data (in kWh) from $300$ individual rooftop PV panels in Australia with $30$ minutes interval available from the Ausgrid dataset
\cite{dataset}. For training purposes, first $9$ months of generation data from house number $2076$ for the year $2011-2012$ is utilized with a validation split of $0.2$ and optimizer $Adam$. The trained model is then tested on remaining $3$ months of energy generation data by making $48$-time steps ahead predictions. The past one day ($48$ intervals) of generation data is passed to the network as historical input to make the algorithm learn.
The results are obtained using python's open source library Tensorflow, $16$ GB RAM, graphics processing unit (GPU), and an $i7$ processor. 
\subsection{Evaluation metrics}
The proposed scheme is evaluated on both deterministic and probabilistic error metrics.  
For Bayesian probabilistic performance, Pinball loss ($Q_l$) and Winkler score ($w_l$) are computed to evaluate the accuracy and sharpness of forecasted quantiles \cite{toubeau2018deep} as:
{\fontsize{9}{10.8}\selectfont
\begin{equation}\label{pinball}
Q_l(y_{t},\hat{y}_{t,\tau},\tau)=\begin{cases}
\tau (y_{t}-{\hat{y}_{t,\tau}})  & y_{t} \ge {\hat{y}_{t,\tau}}\\
(1-\tau)({\hat{y}_{t,\tau}}-y_{t})  & y_{t} \le {\hat{y}_{t,\tau}}
\end{cases}  
\end{equation}}

\noindent where $\hat{y}_{t,\tau}$ symbolizes $t^{th}$ forecast for percentile value $\tau$ against actual solar generation value $y_{t}$.
Furthermore, for confidence level $(1-\gamma) \times 100 $, Winkler score is computed as:
{\fontsize{9}{10.8}\selectfont
\begin{equation}\label{eq:winkler}
W_l=\begin{cases}
\delta & lb_t \le y_t \le ub_t\\
\delta + 2(lb_t-y_t) / \gamma & lb_t \ge y_t\\
\delta + 2 (y_t-ub_t) / \gamma & ub_t \le y_t
\end{cases}
\end{equation}}

\noindent where $lb_t$ and $ub_t$ represent the lower and upper bounds of probabilistic forecasts at $t$, respectively. And, $\delta = ub_t -lb_t $ is defined as the PI width.
A lower Pinball loss and Winkler score imply better probabilistic estimation.

Furthermore, forecasting accuracy at deterministic level is evaluated by computing root-mean square error (RMSE) and mean absolute error (MAE) of the differences between actual generation values and predictive mean as given below:
{\fontsize{9}{10.8}\selectfont
\begin{equation}
RMSE=\sqrt{\frac{1}{n} \sum\limits_{t=1}^{n}(\hat{y}_{t}-y_{t})^2}
\label{rmse}
\end{equation}}
\vspace{-5pt}
{\fontsize{9}{10.8}\selectfont
\begin{equation}
MAE=\frac{1}{n}\sum\limits_{t=1}^{n}\mid{\hat{y}_{t}-y_{t}}\mid
\label{mape}
\end{equation}}
where $n$ denotes the total number of samples in the test data.
\subsection{Comparative analysis}
\begin{figure*}[!t]
	\vspace{-10pt}
	\subfigure[$24$-time steps ahead point predictions]
	{\includegraphics[width=0.35\textwidth]{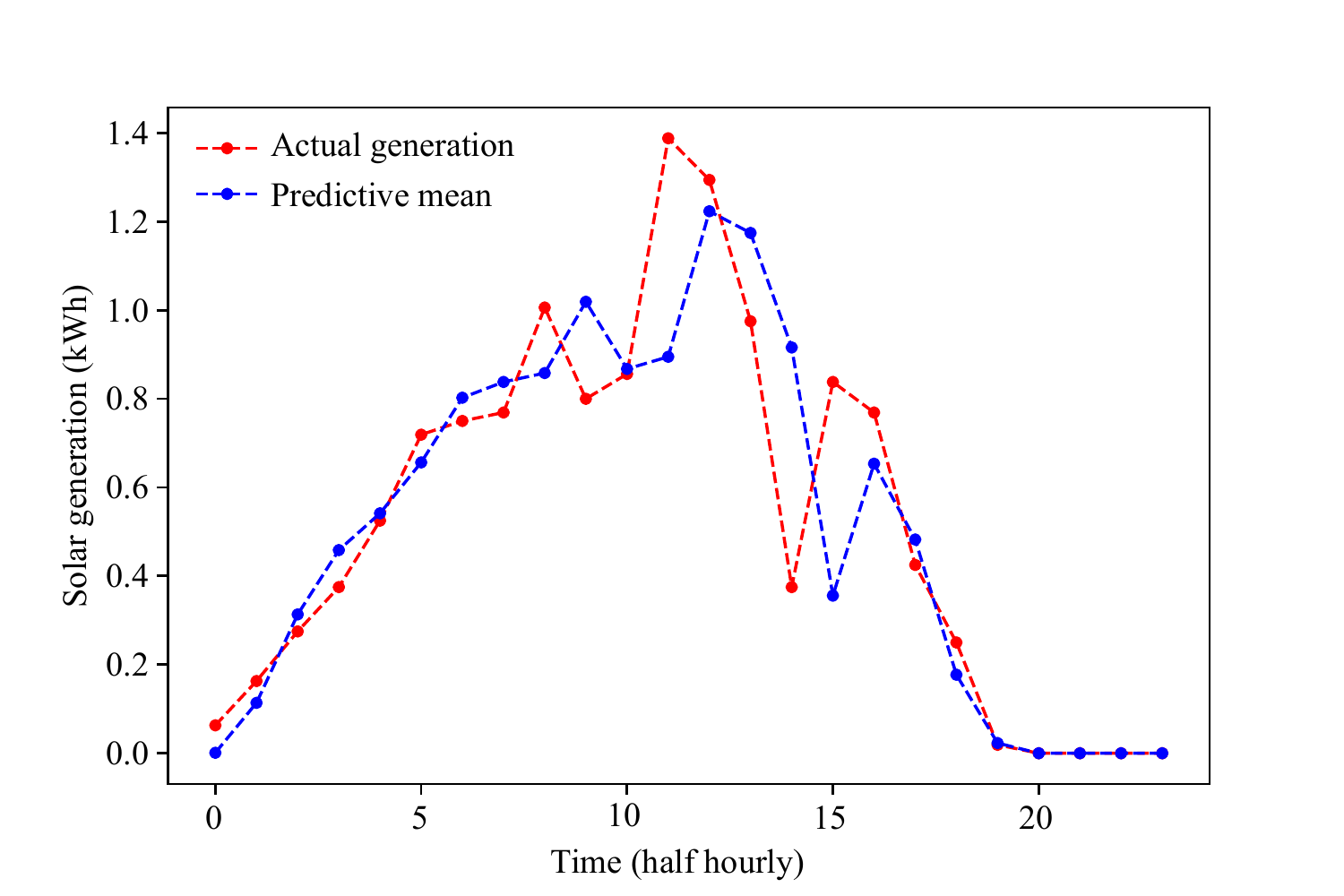}\label{results2a}}	
	\subfigure[PIs for $48$-time steps ahead forecasts]
	{\includegraphics[width=0.35\textwidth]{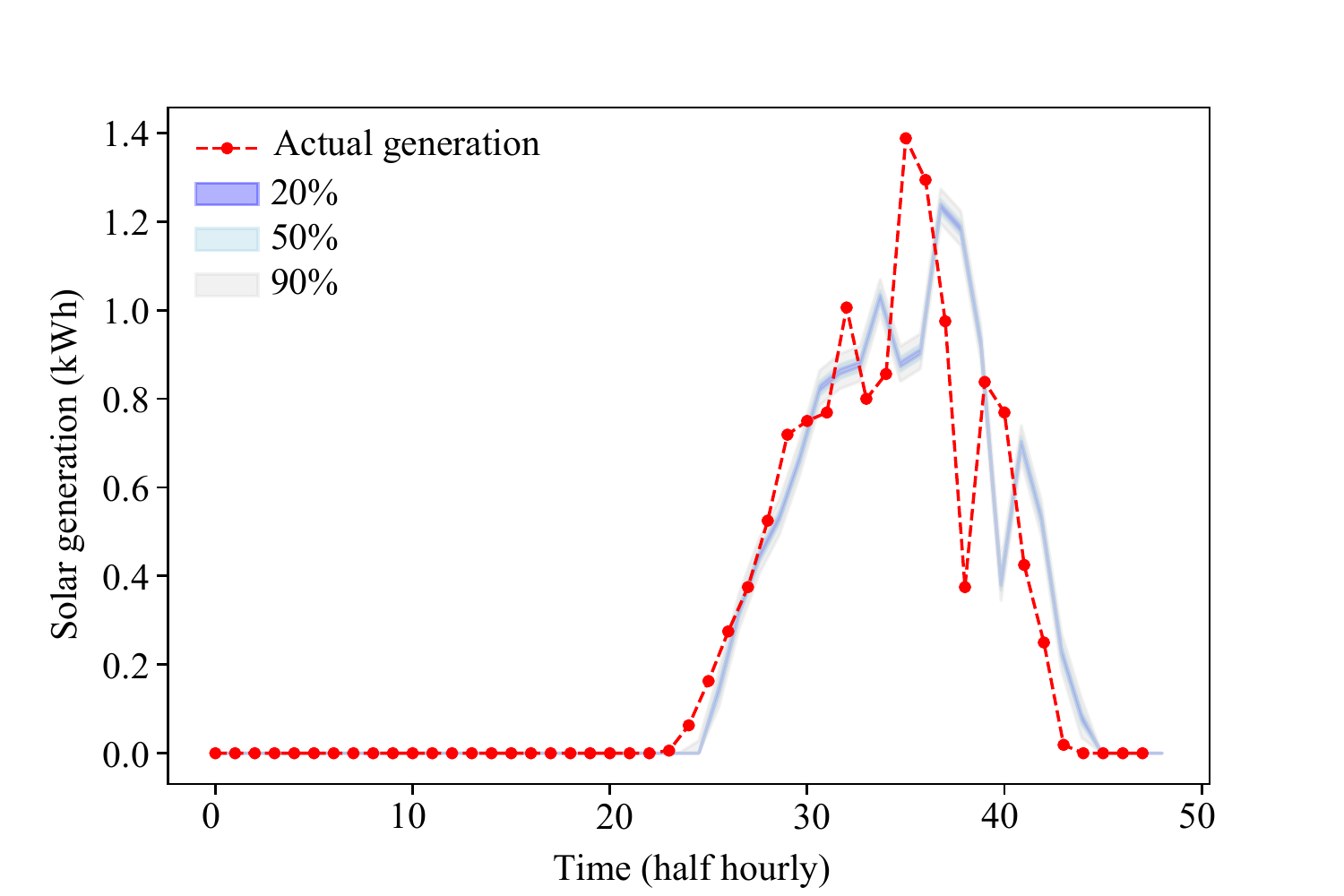}\label{results2b}}
	\subfigure[PIs for one month of data]
	{\includegraphics[width=0.35\textwidth]
		{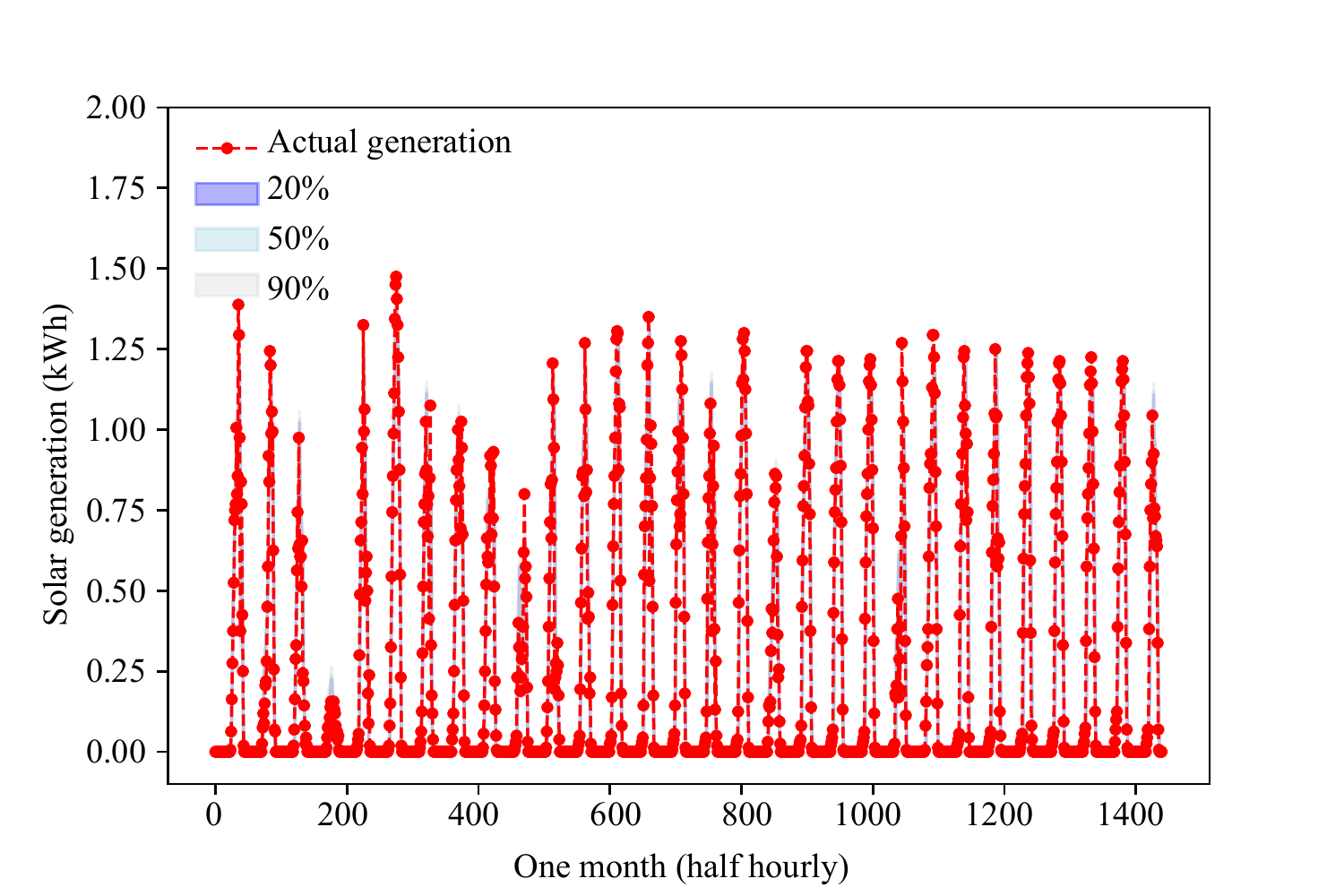}\label{results2c}}
	\caption{Performance analysis of the proposed $Improved$-Bayesian BiLSTM for solar generation forecasting}
	\label{results2}
\end{figure*}
\begin{figure*}[!t]
	\vspace{-10pt}
	\subfigure[$24$-time steps ahead point predictions]
	{\includegraphics[width=0.35\textwidth]{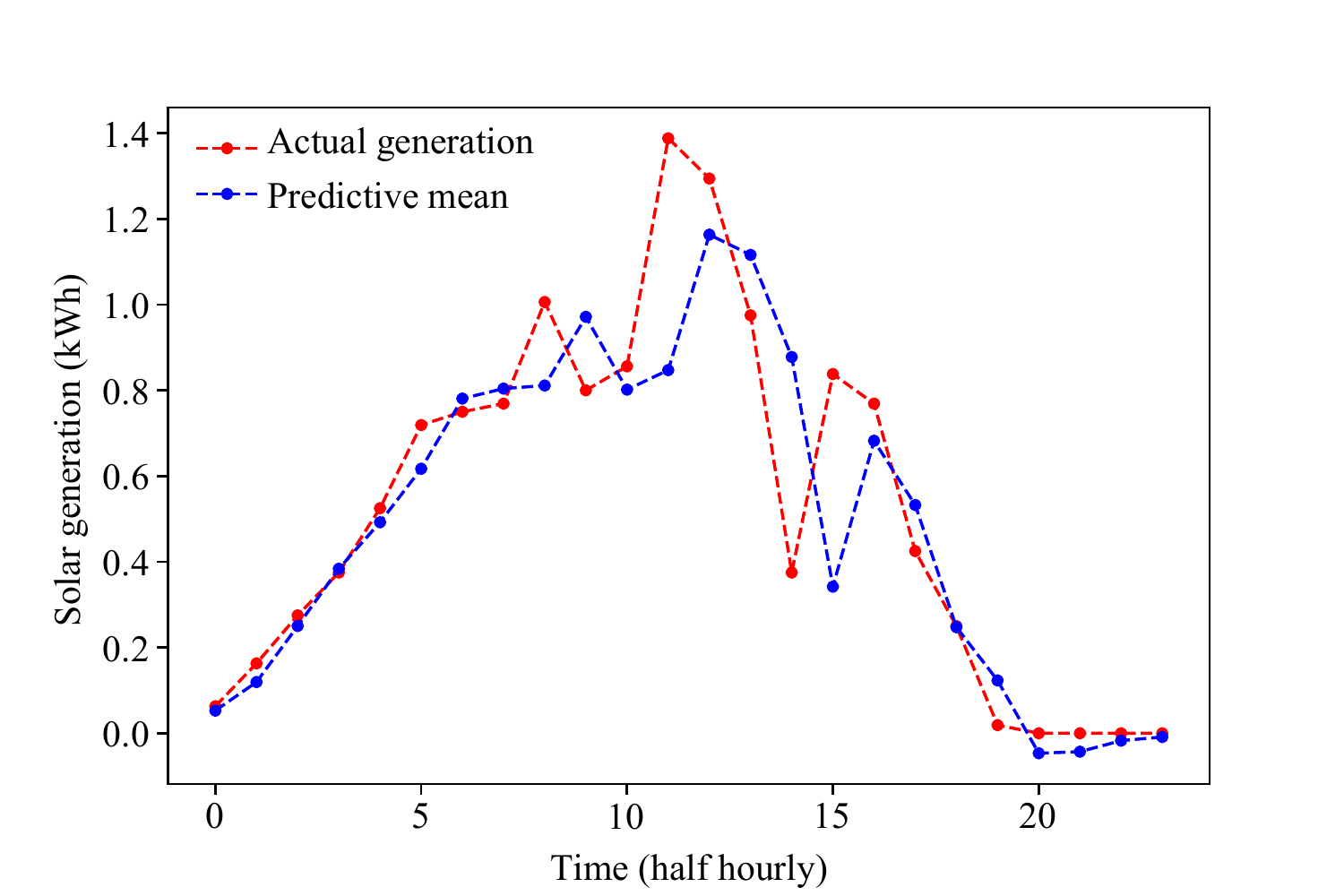}\label{results1a}}	
	\subfigure[PIs for $48$-time steps ahead forecasts]
	{\includegraphics[width=0.35\textwidth]{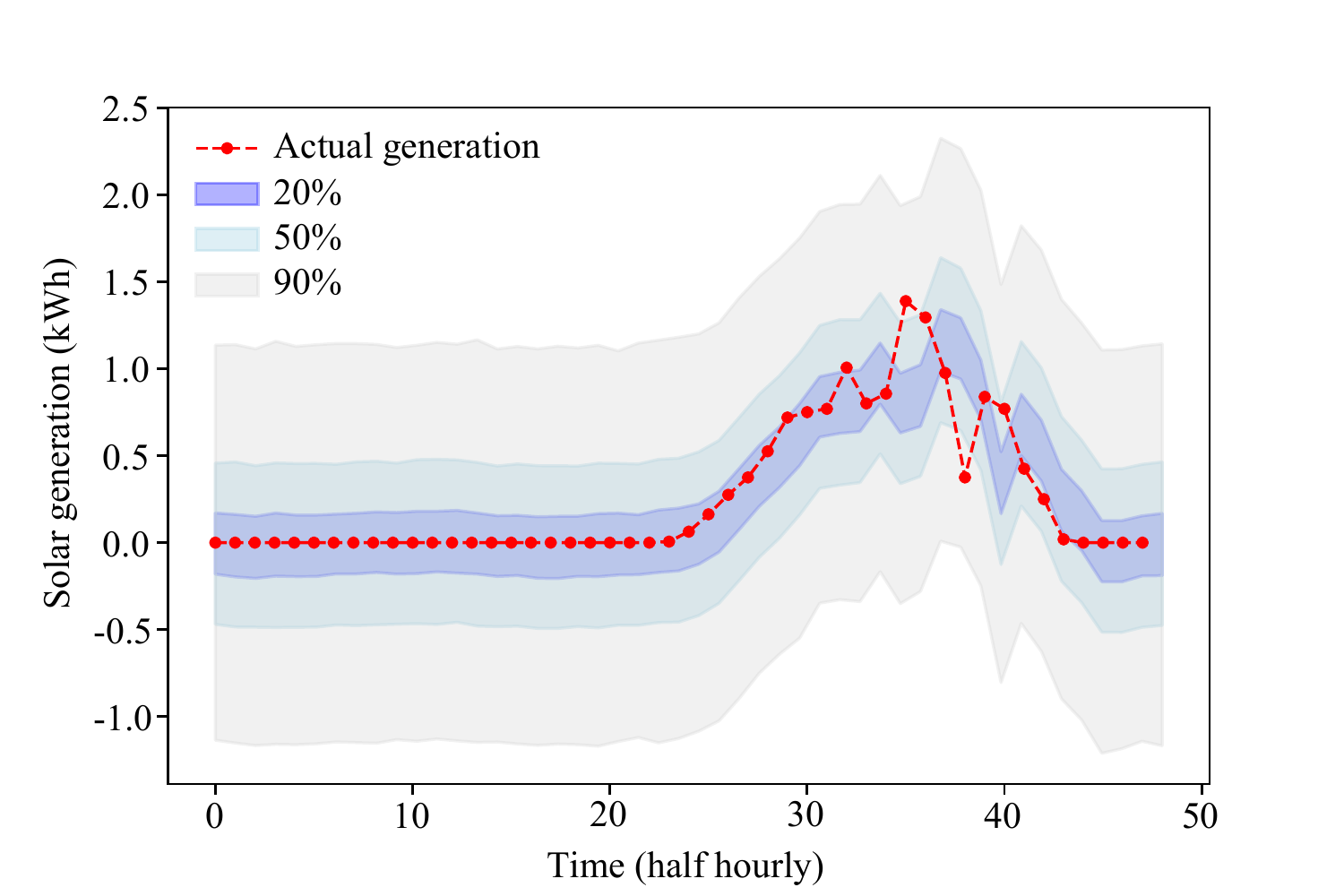}\label{results1b}}
	\subfigure[PIs for one month of data]
	{\includegraphics[width=0.35\textwidth]
	{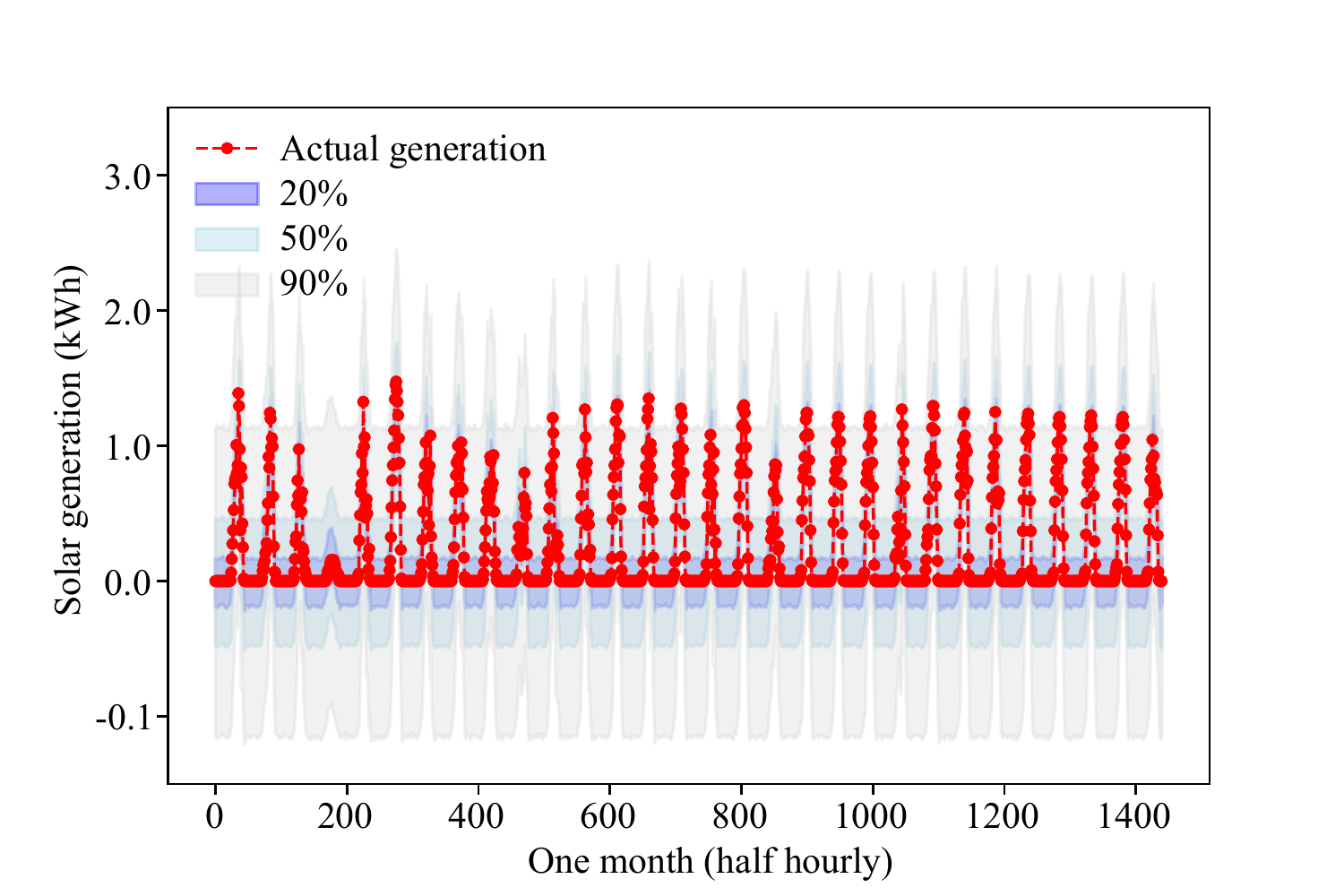}\label{results1c}}
	\caption{Performance analysis of the standard Bayesian BiLSTM for solar generation forecasting}
	\label{results1}
\end{figure*}
To evaluate the forecasting performance, the proposed Bayesian scheme is compared with other variants of standard BDL and benchmark point forecasting methods including RNN, unidirectional LSTM, and BiLSTM architectures. 
The numerical results for the $48$-times steps ahead solar generation forecasting from the proposed and comparative methods are demonstrated in Table \ref{table:msa} using deterministic and probabilistic evaluation metrics.

While point forecasting methods exhibit poor performance for MSA forecasts by demonstrating higher RMSE and MAE values, Bayesian methods with recurrent neural layers, namely Bayesian-RNN, Bayesian-LSTM, and Bayesian-BiLSTM manage to achieve significant decrease in the aforesaid deterministic errors with a competent time complexity. The proposed $Improved$-Bayesian BiLSTM method demonstrates the lowest RMSE and MAE values compared to standard point forecasting and BDL methods.
In addition, standard point forecasting methods are not capable to quantify uncertainty in deep learning models and cannot provide probabilistic forecasts, thus Pinball and Winkler scores are not reported for the first three comparative methods in Table \ref{table:msa}. 

From the probabilistic metrics, it is evident that the proposed $Improved$-Bayesian BiLSTM achieves least Pinball and Winkler scores (in kWh), i.e., $0.0161$ and $0.0143$, respectively which is more than $10$ fold improvement compared to Bayesian-RNN, Bayesian-LSTM, and Bayesian-BiLSTM.
This improvement is entailed by alpha-beta divergence employed in the proposed framework, which can manage uncertainties in the solar generation data and optimize deep learning model parameters more effectively. On the other hand, standard Bayesian BiLSTM uses KL divergence which underestimates the variance of the approximated posterior.
Also, it is important to note that BiLSTM with point and probabilistic approach takes additional execution time due to bidirectional processing involved in the learning process. However, the $Improved$-Bayesian BiLSTM requires less computation time compared to standard Bayesian BiLSTM as the solutions can converge effectively when the weight parameters are optimized using a more accurate divergence technique.  

Furthermore, Fig. \ref{results2} and Fig. \ref{results1} represent graphical illustrations for solar generation forecasts using $Improved$ and standard Bayesian BiLSTM, respectively. Figs. \ref{results2a} and \ref{results1a} reflect similar point forecasting patterns between actual and predictive mean with a time lag during the peak irradiance hours. On the other hand, Fig. \ref{results2b} and Fig. \ref{results1b} highlight the probabilistic forecasts for next day ahead ($48$-time steps ahead) solar generation with $20\%$, $50\%$, and $90\%$ PIs using $Improved$-Bayesian and standard Bayesian BiLSTM method, respectively. 
Given that the $Improved$-Bayesian BiLSTM estimates a more accurate approximation of the posterior distribution of weight parameters, Fig. \ref{results2b} demonstrates that the proposed method improves the future generation probabilities for different percentiles, by providing tighter PI bounds. 
Furthermore, Fig. \ref{results2c} and Fig. \ref{results1c} provide MSA probabilistic forecasts on one month of data using both methods ensuring the capability of proposed method to quantify uncertainty more precisely for longer horizons.

\begin{table}[!ht]
	\renewcommand{\arraystretch}{1.4}
	\vspace{-8pt}
	\caption{MSA probabilistic forecasting results for ($t+24$) }\label{table:t+24}
	\begin{tabular}{p{25mm}p{8mm}p{8mm}p{8mm}p{8mm}}
		\hline
		Method &RMSE &MAE & Pinball&Winkler\\
		\hline				
		Bayesian BiLSTM&$0.0963$&$0.0449$&$0.2770$&$1.7050$\\
		$Improved$-Bayesian BiLSTM&$0.0854$&$0.0370$&$0.0148$&$0.0123$\\
		\hline
	\end{tabular}
\end{table}
\begin{table}[!ht]	
	\renewcommand{\arraystretch}{1.4}
	\vspace{-8pt}
	\caption{MSA probabilistic forecasting results for ($t+12$) }\label{table:t+12}
	\begin{tabular}{p{25mm}p{8mm}p{8mm}p{8mm}p{8mm}}
		\hline
		Method&RMSE &MAE & Pinball &Winkler\\
		\hline				
		Bayesian BiLSTM&$0.0954$&$ 0.0475$&$0.2653$&$1.6170$\\
		$Improved$-Bayesian BiLSTM&$0.0851$&$0.0357$&$0.0138$&$0.0121$\\
		\hline
	\end{tabular}
\end{table}
\begin{table}[!ht]	
	\renewcommand{\arraystretch}{1.4}
	\vspace{-8pt}
	\caption{SSA probabilistic forecasting results for ($t+1$) }\label{table:t+1}
	\begin{tabular}{p{25mm}p{8mm}p{8mm}p{8mm}p{8mm}}
		\hline
		Method &RMSE &MAE & Pinball &Winkler\\
		\hline		
		Bayesian BiLSTM&$0.0928$&$0.0469$&$0.1446$&$1.4170$\\
		$Improved$-Bayesian BiLSTM&$0.0850$&$0.0350$&$0.0127$&$0.0115$\\
		\hline
	\end{tabular}
\end{table}

We further extend the experiments to three more variations involving 24-, 12-, and 1-time steps ahead generation forecasts conducted on standard Bayesian BiLSTM and the proposed $Improved$-Bayesian BiLSTM.
Tables \ref{table:t+24}, \ref{table:t+12}, and \ref{table:t+1} illustrate numerical results for respective variations using deterministic and probabilistic forecasting evaluation metrics. 
It can be observed that RMSE and MAE values for the proposed method are lower than those of the Bayesian BiLSTM method from all three cases. In addition, $improved$ technique achieves a more than $10$-fold improvement in the Pinball and Winkler scores for all three variations of MSA solar generation forecasting. Also, it is worthwhile to observe that with the decrease in forecasting time horizons, i.e., from $48$ to SSA ($30$ minutes ahead), the forecasting accuracy increases in terms of lower error values. To be specific, there is a $17\%$ ($50\%$) degradation in the Winkler score (Pinball loss) for the standard Bayesian BiLSTM method when forecasts are generated $48$-time steps ahead rather than $1$-time step ahead. On the other hand, the $improved$ method suffers a degradation of only $6.5\%$ ($14\%$) in the Winkler score (Pinball loss). Thus, the proposed method is more suitable for generating MSA forecasts as compared to standard Bayesian BiLSTM.
\begin{table}[!h]
	\centering
		\renewcommand{\arraystretch}{1.2}
		\vspace{-8pt}
	\caption{Impact of alpha-beta parameters on $Improved$-Bayesian BiLSTM $(t+48)$}\label{table:alphabeta}
	\begin{tabular}{p{5mm}p{5mm}p{8mm}p{8mm}}
		\hline
		$\alpha$&$\beta$&Pinball&Winkler \\
		\hline
		$1.0$&$2.0$&$0.0161$&$0.0143$\\	
	    $2.0$ & $3.0$ & $0.0190$&$0.0210$\\
		$3.0$ & $4.0$ &$0.0224$&$0.0514$\\
		$2.0$ & $6.0$ &$0.0422$&$0.2085$\\
		$3.0$ & $8.0$ &$0.0638$&$0.2663$\\
		$5.0$ & $15.0$ &$0.1143$&$0.2805$\\
		\hline
	\end{tabular}
\end{table}

Table \ref{table:alphabeta} represents the impact of different $\alpha$ and $\beta$ parameters on Pinball and Winkler scores for $48$-time steps ahead solar generation forecasting.
It can be observed from the experiments that lower values of alpha and beta achieve lower errors, and thus provide better forecasting performance. 
\begin{figure}[!h]
\vspace{-10pt}
	\centering
	\includegraphics[width=0.4\textwidth]{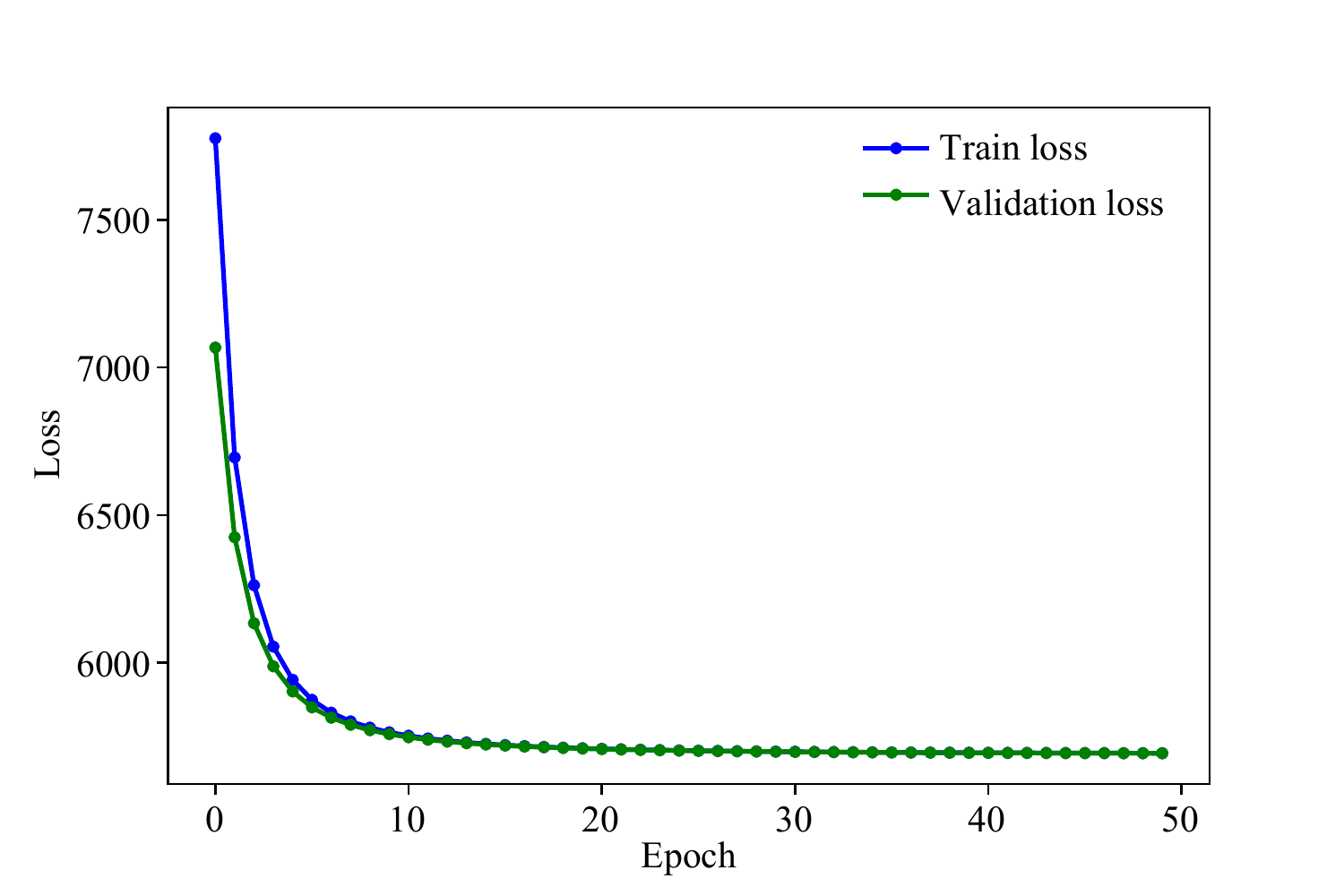}
	\vspace{-10pt}
	\caption{Loss reduction using standard Bayesian BiLSTM}
	\label{loss1}
\end{figure}
\begin{figure}[!h]
\vspace{-10pt}
	\centering
	\includegraphics[width=0.4\textwidth]{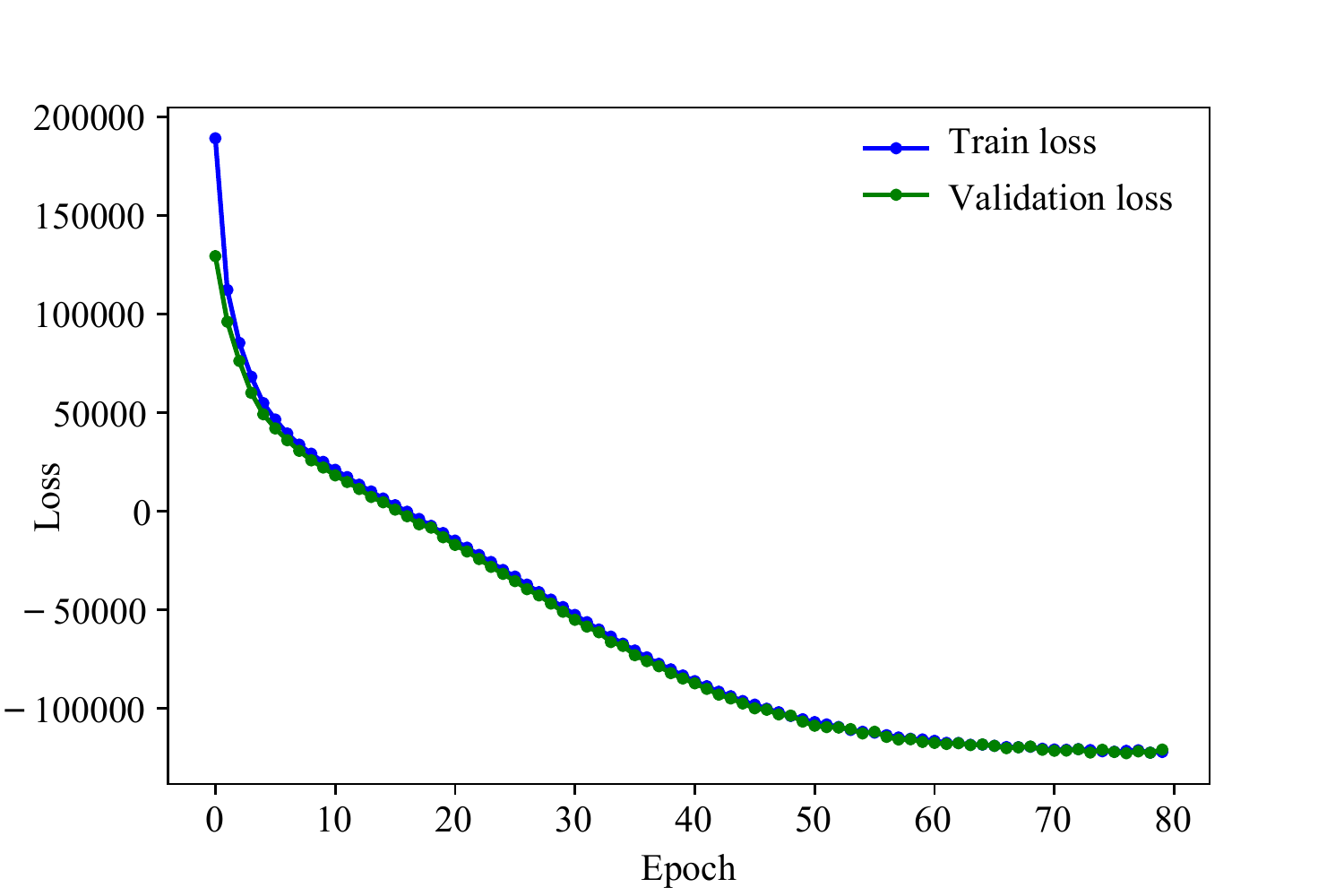}
	\vspace{-10pt}
	\caption{Loss reduction using $Improved$-Bayesian BiLSTM}
	\label{loss2}
\end{figure}

Furthermore, Fig. \ref{loss1} and Fig. \ref{loss2} illustrate the convergence of likelihood loss against number of epochs validated with $0.2$ validation split during the network training phase. The aforementioned figures demonstrate effective learning process via Bayesian technique without overfitting or underfitting the training data. However, the proposed $Improved$-Bayesian BiLSTM achieves lower loss as compared to standard Bayesian method as number of epochs increase. This is because the loss function employed in the learning process is negative log likelihood which is more accurately treated by the alpha-beta divergence technique. 
\section{Conclusion}\label{sec:conclusion}
In this paper, an improved MSA forecasting approach is presented using Bayesian probability integrated with alpha-beta divergence and BiLSTM deep learning method to predict future energy generation in smart grid systems. The proposed method is evaluated using solar generation data and forecasting results demonstrate the efficacy of point and probabilistic predictions in the terms of lower error values. Furthermore, impact of different time horizons on the forecasting performance of the proposed method has also been investigated. The proposed method outperforms standard Bayesian techniques for both SSA and MSA forecasting by demonstrating least prediction errors with less computational time. Future work will focus on exploring Bayesian BiLSTM for multidimensional energy forecasting problems.
\bibliographystyle{IEEEtran}
\appendix
This appendix presents the derivation for alpha-beta divergence ($D_{\alpha\beta}$) given in the Eq. \eqref{eq:finalalphabeta} as an extension to the following scale-invariant alpha-beta divergence from \cite{regli2018alpha}:
{\fontsize{9}{10.8}\selectfont
\begin{multline}
D_{\alpha\beta}(q||p) =\frac{1}{\alpha (\alpha + \beta)} \log \mathbb{E} [\frac{p(\theta,w)^ {\alpha + \beta}}{q(\theta)}]  \\
+ 
\frac{1}{\beta (\alpha + \beta)} \log \mathbb{E} [q(\theta)^{\alpha+\beta-1}]\\
-
\frac{1}{\alpha\beta} \log \mathbb{E} [q(\theta)^{\alpha+\beta-1} \left(\frac{p(\theta,w)}{q(\theta)}\right)^{\beta}] 
\end{multline}}
The aforementioned expression can be written as: 
{\fontsize{9}{10.8}\selectfont
\begin{align}
D_{\alpha\beta}(q||p) & \ge
\frac{1}{\alpha (\alpha + \beta)} \mathbb{E}  [\log \frac{p(\theta,w)^ {\alpha + \beta}}{q(\theta)}]  \nonumber\\
&+ 
\frac{1}{\beta (\alpha + \beta)} \mathbb{E} [\log [q(\theta)^{\alpha+\beta-1}]\nonumber\\
&-
\frac{1}{\alpha\beta} \mathbb{E} [\log q(\theta)^{\alpha+\beta-1} \left(\frac{p(\theta,w)}{q(\theta)}\right)^{\beta}] \label{eq:2}\\
& =
\frac{1}{\alpha (\alpha + \beta)} \mathbb{E}  [\log {p(\theta,w)^ {\alpha + \beta}}-\log {q(\theta)}] \nonumber\\
&+ 
\frac{1}{\beta (\alpha + \beta)} * (\alpha+\beta-1) * \mathbb{E} [\log q(\theta)]\nonumber\\
&-
\frac{1}{\alpha\beta} \mathbb{E} [\log q(\theta)^{\alpha+\beta-1} ]
-\frac{1}{\alpha\beta} \log \left(\frac{p(\theta,w)}{q(\theta)}\right)^{\beta} \label{eq:3}\\
& =
\frac{1}{\alpha} \mathbb{E}  [\log {p(\theta,w)}]-\frac{1}{\alpha (\alpha + \beta)}\log [{q(\theta)}] \nonumber\\
&+ 
\frac{\alpha+\beta-1}{\beta (\alpha + \beta)}  \mathbb{E} [\log q(\theta)]-
\frac{\alpha+\beta-1}{\alpha\beta} \mathbb{E} [\log q(\theta)]\nonumber\\
&-\frac{1}{\alpha} \mathbb{E}[\log p(\theta,w)]
+ \frac{1}{\alpha}\log {q(\theta)} \label{eq:4}\\
& =
\left(\frac{\alpha+\beta-1}{\beta(\alpha+\beta)} - \frac{\alpha + \beta -1}{\alpha\beta}-\frac{1}{\alpha (\alpha+\beta)} + \frac{1}{\alpha} \right) \nonumber\\
&\times \mathbb{E}[\log q(\theta)]\label{eq:5}
\end{align}}

\noindent where \eqref{eq:2} follows Jensen's inequality and \eqref{eq:3}-\eqref{eq:4} are derived using logarithmic operations. Finally, \eqref{eq:5} is written after some mathematical manipulations.


\end{document}